\documentclass[a4paper, twoside]{article}
\usepackage{circuitikz}
\usepackage{todonotes}
\usepackage{epsfig}
\usepackage{subfigure}
\usepackage{calc}
\usepackage{natbib}
\let\bibhang\relax
\usepackage{amssymb}
\usepackage{amstext}
\usepackage{amsmath}
\usepackage{algorithm}
\usepackage{algpseudocode}
\usepackage{amsthm}
\usepackage{hyperref}
\usepackage{multicol}
\usepackage{url}
\usepackage{pslatex}
\usepackage{csvsimple}
\usepackage{tikz}
\usepackage{listings}
\usepackage{svg}
\usepackage{float}
\usepackage{apalike}
\usepackage{balance}
\usepackage[bottom]{footmisc}
\usetikzlibrary{shapes,external}
\usepackage{balance}
\usepackage{SCITEPRESS}

\newcommand*{\wf}{\mathcal{W}}
\newcommand*{\txt}{\mathcal{T}}
\newcommand*{\acts}[1]{\mathcal{A}(#1)}

\theoremstyle{definition}
\newtheorem*{defi}{Definition}

\begin{document}

\title{GUIDO: A Hybrid Approach to Guideline Discovery \& Ordering from Natural Language Texts}
\author{\authorname{Nils Freyer\sup{1}\orcidAuthor{0000-0002-4460-3650}, Dustin Thewes\sup{1}        \orcidAuthor{0000-0002-1301-8926} and Matthias Meinecke\sup{1}\orcidAuthor{0009-0008-3055-5505}}       \affiliation{\sup{1}FB7 Operations Management, FH Aachen University of Applied Sciences, Aachen, Germany}       \email{\{freyer, thewes, meinecke\}@fh-aachen.de}
}


\abstract{Extracting workflow nets from textual descriptions can be used to simplify guidelines or formalize textual descriptions of formal processes like business processes and algorithms. The task of manually extracting processes, however, requires domain expertise and effort. While automatic process model extraction is desirable, annotating texts with formalized process models is expensive. Therefore, there are only a few machine-learning-based extraction approaches. Rule-based approaches, in turn, require domain specificity to work well and can rarely distinguish relevant and irrelevant information in textual descriptions. In this paper, we present GUIDO, a hybrid approach to the process model extraction task that first, classifies sentences regarding their relevance to the process model, using a BERT-based sentence classifier, and second, extracts a process model from the sentences classified as relevant, using dependency parsing. The presented approach achieves significantly better results than a pure rule-based approach. GUIDO achieves an average behavioral similarity score of $0.93$. Still, in comparison to purely machine-learning-based approaches, the annotation costs stay low.}

\onecolumn \maketitle \normalsize \setcounter{footnote}{0} \vfill

\section{\uppercase{Introduction}}
To fulfill a task or execute a process in a predetermined way, especially when lacking the respective expertise, one often needs to follow \textit{guidelines}. Guidelines are commonly given as unstructured texts. Examples from their domain space are business processes, technical standards, cooking recipes, medical guidelines explaining the standard procedures to medical professionals, or the description of algorithms. Understanding, updating, and conformance-checking a guideline requires sufficient proficiency in the language, adequate reading comprehension, and often adequate domain expertise (e.g., a medical degree). 

In contrast to unstructured texts, process models may be described using formalized process modeling. Process models encode order, decision rules, and loops in the notation, only requiring labeling of the activities, constraints, and decision rules as texts \citep{semantic_challenges_mendling_2014}. However, transforming unstructured text into structured process models requires expertise in process modeling and thus, yields an expensive task \citep{king_process_2011,frederiks_information_2006}.

The assisted extraction of formalized process models from text is an active field of research and could alleviate those problems \citep{lopez_assisted_2019}.  Contemporary approaches are either pure rule-based, usually specific to a domain, or purely machine-learning-based, requiring large amounts of annotated data for a specific domain and language. As extracting process models manually is time-consuming and expensive, using pure machine-learning-based approaches is either restricted to domains with a sufficient amount of annotated data or requires large corpora to be annotated, making it inapplicable for smaller extraction domains.

We propose GUIDO, a Guideline Discovery \& Ordering approach that extracts process models from natural language text (cf. \autoref{sec:guido}). GUIDO first uses a BERT sequence classifier to identify and filter sentences relevant to the process. Second, it uses a language rule-based model to extract the processes' activities, interactivity relations, and temporal order. Finally, GUIDO uses the extracted relations to formalize the process model as a workflow net. 
We demonstrate the proposed approach with German recipes, achieving an F1-score of $0.973$ for sentence classification and an average behavioral similarity score between generated process models and human-expert-made process models of $0.93$ (cf. \autoref{sec:experiments}). 
The code and data for this project are publicly available at \url{https://github.com/nils-freyer/GUIDO}

\section{\uppercase{Ethical Considerations}}
While this paper investigates extracting process models on German recipes, the approach applies to a more extensive section of the domain space, including more safety- and security-relevant domains. The approach introduced in this paper merely offers assistance in extracting process models. Both the rule-based component and the machine-learning-based component of the approach may not generalize to use cases outside the evaluation scenario. Furthermore, pre-trained BERT models will introduce biases to the text classification \citep{liang2021towards}. Depending on the application domain, discriminatory outcomes should be examined carefully.

\section{\uppercase{Related Work}}
Process Model Extraction (PME) is considered a Text to Model challenge, including identifying activities and their sequence or concurrency \citep{semantic_challenges_mendling_2014}. PME approaches can be categorized broadly as \textit{rule-based}, 
\textit{machine-learning-based},
or \textit{hybrid}, combining rule and machine-learning-based approaches.

\paragraph{Rule-Based Approaches.} Rule-based approaches mainly use grammatical features of a text and are applied to both extracting declarative \citep{aa2019extracting,winter2018detecting} and imperative \citep{zhang2012automatically,walter2011workflow,schumacher2012extraction} process models. Although they perform domain-specifically well, restrictions have to be made to identify activities as, e.g., verb centrality \citep{walter2011workflow,qian2020approach} or constraint markers \citep{aa2019extracting,winter2018detecting,winter2019deriving} requiring domain-specific knowledge on potential heuristics. 

\paragraph{Machine-Learning-Based Approaches.}
Machine-learning-based approaches such as conditional random fields, support vector machines, and neural text classification was used for the detection of the process relevant sentences \citep{leopold2018identifying,qian2020approach}. 
Furthermore, Qian et al \citep{qian2020approach} identified process model extra as a multi-grained text classification task. They developed a hierarchical neural network to classify relevant sentences and generate the extracted process model. While the results are promising, a multi-grained, annotated dataset is needed.
Additionally, to the related task of extracting linear temporal logic from natural language texts, a neural machine translation approach was proposed \citep{brunello2019synthesis}.

\paragraph{Hybrid Approaches.}
Little work has combined rule-based and machine-learning-based PME approaches. Relatedly, \citet{winter2019deriving} used constraint markers as \textit{shall, must, should}, to identify sentences containing declarative process information and used sentence embeddings and clusterings to find related constraints. However, these examples do not implement hybrid approaches for the extraction of process models.
\\ \ \\
To the best of our knowledge, there were no implementations and evaluation on German texts yet. Especially rule-based approaches will differ language-wise.
Furthermore, GUIDO is the first hybrid PME approach, using generally known approaches in a novel hybrid way in order to reduce labeling costs and maximize generalizability and accuracy.

\section{\uppercase{GUIDO as a multi-level extraction model}}
\label{sec:guido}
As described by \citet{qian2020approach}, the PME task can be formulated as a hierarchical information extraction task. That is, we can subdivide the task into sentence classification, activity extraction and activity ordering.
This section introduces basic preliminaries, notations and outlines the proposed solutions to each of the sub-tasks. 
\subsection{\uppercase{Preliminaries}}    
Within our research, we chose to use Petri nets \citep{chen1990petri} and more specifically workflow nets \citep{van1998application} to formalize imperative process models. 
\begin{defi}[Workflow Net]
    A Petri net is a tuple $N=(P,T,F)$, where $P$ is a set of places, $T$ is a set of transitions, $P \cap T = \emptyset$, and $F \subseteq (P \times T) \cup (T \times P)$ is the flow relation of the network.
\end{defi}
A workflow net is a Petri net $\wf=(P,T,F)$, such that there is a unique source and a unique sink to all paths in the net. Especially in our domain, workflow nets, as a subclass of Petri nets, are a reasonable choice, as any recipe has a dedicated set of end states and thus, can be converted to a workflow  net. The transitions of the Petri net describe the activities of the process. An activity is typically constituted by the act (verb), its subjects and objects, as well as its modifiers.
\begin{defi}[Activity]
    Given a vocabulary $V$, an activity is a tuple $a = (v, s, o, m) \in \mathcal{P}(V)^4$, where $v$ is a set of verbs, s is a set of subjects, $o$ is a set of objects and $m$ is a set of modifiers declaring the activity. Given a text $\txt=(S_1,\dots,S_n)$ with sentences $S_1, \dots, S_n \in V^{m}$, $m \in \mathfrak{N}$, $\acts{\txt}$ denotes the set of activities in $\txt$ and consequently $\acts{S}$ denotes the set of activities in a given sentence $S$. 
\end{defi}
For instance $(\text{"foam"}, \text{"butter"}, \emptyset, \text{"in a hot pan"})$, is the activity we want to extract from the sentence "Foam butter in a hot pan".
Therefore, if we want to extract a workflow net $\wf$ from a text $\txt$ we derive the following extraction task. 
\begin{defi}[Process Model Extraction Task]
\label{pmet}
    Given a text $\txt=(S_1,\dots,S_n)$, extract a workflow net $N=(P,T,F)$, s.t. $T = \acts{\txt}$ and $F$ spans the temporal relation of $\acts{\txt}$ in $\txt$.
\end{defi}
\subsection{\uppercase{Model Architecture}}
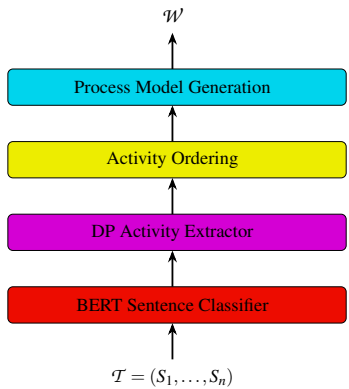
\begin{figure}[h]
\centering
\resizebox{0.6\linewidth}{!}{%
\begin{circuitikz}
    \tikzstyle{every node}=[font=\LARGE]
    
    \draw [ fill={rgb,255:red,237; green,12; blue,0} , rounded corners, ] (8.5,3.5) rectangle (17.75,2.5);
    \draw [line width=0.5mm, -Stealth] (13,3.5) -- (13,4.5);
    \draw [ fill={rgb,255:red,212; green,0; blue,212} , rounded corners, ] (8.5,5.5) rectangle (17.75,4.5);
    \draw [line width=0.5mm, -Stealth] (13,5.5) -- (13,6.5);
    \draw [ fill={rgb,255:red,237; green,237; blue,0} , rounded corners, ] (8.5,6.5) rectangle (17.75,7.5);
    \draw [ fill={rgb,255:red,0; green,212; blue,237} , rounded corners, ] (8.5,8.5) rectangle (17.75,9.5);
    \node [font=\LARGE] at (13,1) {$\txt =(S_1,\dots,S_n)$};
    \draw [line width=0.5mm, -Stealth] (13,1.5) -- (13,2.5);
    \node [font=\LARGE] at (13,3) {BERT Sentence Classifier};
    \node [font=\LARGE] at (13,5) {DP Activity Extractor};
    \node [font=\LARGE] at (13,7) {Activity Ordering};
    \draw [line width=0.5mm, -Stealth] (13,7.5) -- (13,8.5);
    \node [font=\LARGE] at (13,9) {Process Model Generation};
    \draw [line width=0.5mm, -Stealth] (13,9.5) -- (13,10.5);
    \node [font=\LARGE] at (13,11) {$\wf$};
\end{circuitikz}
}%
\caption{Hierarchical model architecture}
\label{fig:arc}
\end{figure}

Understanding PME task as a hierarchical information extraction task, first, we need to classify whether a particular sentence $S$ of a text $\txt$ contains an activity $a \in \acts{\txt}$. Second, we need to extract all $a_1,\dots, a_k$ in $\acts{S}$. Finally, we need to extract the temporal order $\mathrm{T}$ of $\acts{\txt}$ (cf. \autoref{fig:arc}), in order to derive the flow relation $F$ of the workflow net.

Each sub-task was implemented and evaluated separately in addition to the total evaluation of the extracted workflow nets. Therefore, they can be used independently to create baselines for the hybrid approach. 

\subsection{Sentence Classifiers}
The sentence classification level of GUIDO has to perform the binary classification task $\acts{S} = \emptyset$,  given a sentence $S$ in a text $\txt$, i.e., whether a sentence contains an activity or not. We implemented and tested three different classification strategies and compared them to a rule-based baseline strategy.
\paragraph{VVIMP Rule-Based Baseline.}
As a rule-based approach, we implemented a heuristic that classifies a sentence as process relevant if there is no subject that is not a child of an imperative in the dependency tree.
\paragraph{LSTM Classifier.}
A simple LSTM \citep{hochreiter1997long} with a text-classification head was implemented and fully configured by hydra configurations. The LSTM was optimized by a hyper parameter search with 5 workers. The documents were vectorized using either pre-trained and fine tuned GloVe\footnote{Pre-trained glove vectors taken from: \url{https://www.deepset.ai/german-word-embeddings}} vectors or pre-trained FastTexts\footnote{Pre-trained FastText vectors taken from: \url{https://fasttext.cc/docs/en/crawl-vectors.html}} vectors.
\paragraph{Logistic Regression.}
A binary logistic regression classifier was implemented using tfidf document vectorization. 
\paragraph{BERT Sequence Classifier}
The huggingface's BERT \citep{devlin_bert_2019} for sequence classification was used\footnote{\url{https://huggingface.co/docs/transformers/v4.26.0/en/model_doc/bert\#transformers.TFBertForSequenceClassification}}, using a linear layer for classification on the pooled output of the BERT model. The pre-trained German BERT transformer model \citep{chan2023german} was  used to initialize the model. The German BERT model was chosen over the multilingual pre-trained BERT, as it has shown superior performance on common evaluation sets \citep{chan2023german}.

\subsection{Activity Extraction by Dependency Grammar}
\label{subsub:rel}
The next level of GUIDO performs the task of activity extraction.
Given a sentence $S$ with $\acts{S} \neq \emptyset$, we want to extract all activity relations $a_1,\dots,a_n \in \acts{S}$. Machine-learning-based relation extraction models require complexly annotated corpora. Therefore, to reduce annotation costs, we chose to implement a rule-based relation extraction approach, using dependency grammar \citep{nivre2005dependency}.
Dependency grammar is a school of grammar that describes the hierarchical structure of sentences based on dependencies between words within a sentence. NLP frameworks such as spaCy have incorporated dependency parsers into their pipelines \citep{honnibal2020spacy}, trained on large news corpora. Thus, using dependency parsers, POS tags, and STTS tags \citep{albert2003tiger}, does not require further manual labeling.
Dependency grammar-based approaches were proposed to be used for the extraction of process activities from text \citep{sintoris2017extracting,kolb2013creating,zhang2012automatically} as well as for similar tasks such as the translation of sentences to linear temporal logic \citep{brunello2019synthesis} or the extraction of declarative process constraints from natural language texts \citep{winter2018detecting,aa2019extracting}. A major pitfall of using a dependency grammar for activity extraction are non-relevant sentences and subordinate clauses. Therefore, it was primarily applied to documents with strict language norms, e.g., laws, where rule-based classifiers, taking markers as \textit{must} or \textit{should} as indicators of a relevant sentence, work particularly well.  
As we use a sentence classifier to avoid irrelevant sentences, handling subordinate clauses remains on the activity extraction level of the PME task.
\paragraph{Extraction Rules.}
\label{par:er}
By assumption, we extract activities from relevant sentences only. Therefore, activities are expressed as verbs with dependent subjects, objects, and modifiers. In rare cases, activities may be expressed as passivized subjects\citep{aa2019extracting}.

\begin{figure}[h]
    \centering
\includegraphics[trim=0cm 0cm 0cm 5cm, width=\linewidth]{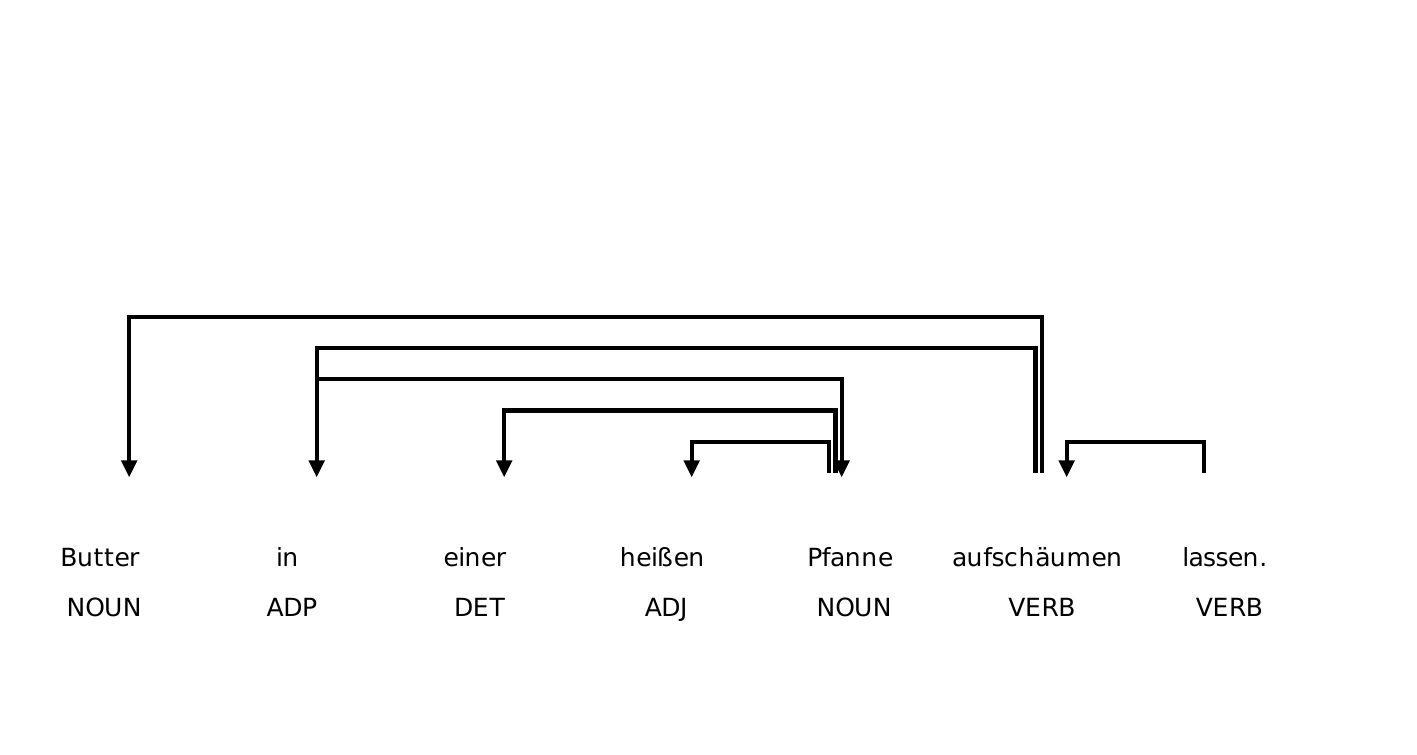}
    \caption{Dependency tree of a German recipe sentence}
    \label{fig:dp_sentence}
\end{figure}

\begin{figure}[ht]
    \centering
    \includegraphics[trim=0cm 0cm 0cm 5cm, width=\linewidth]{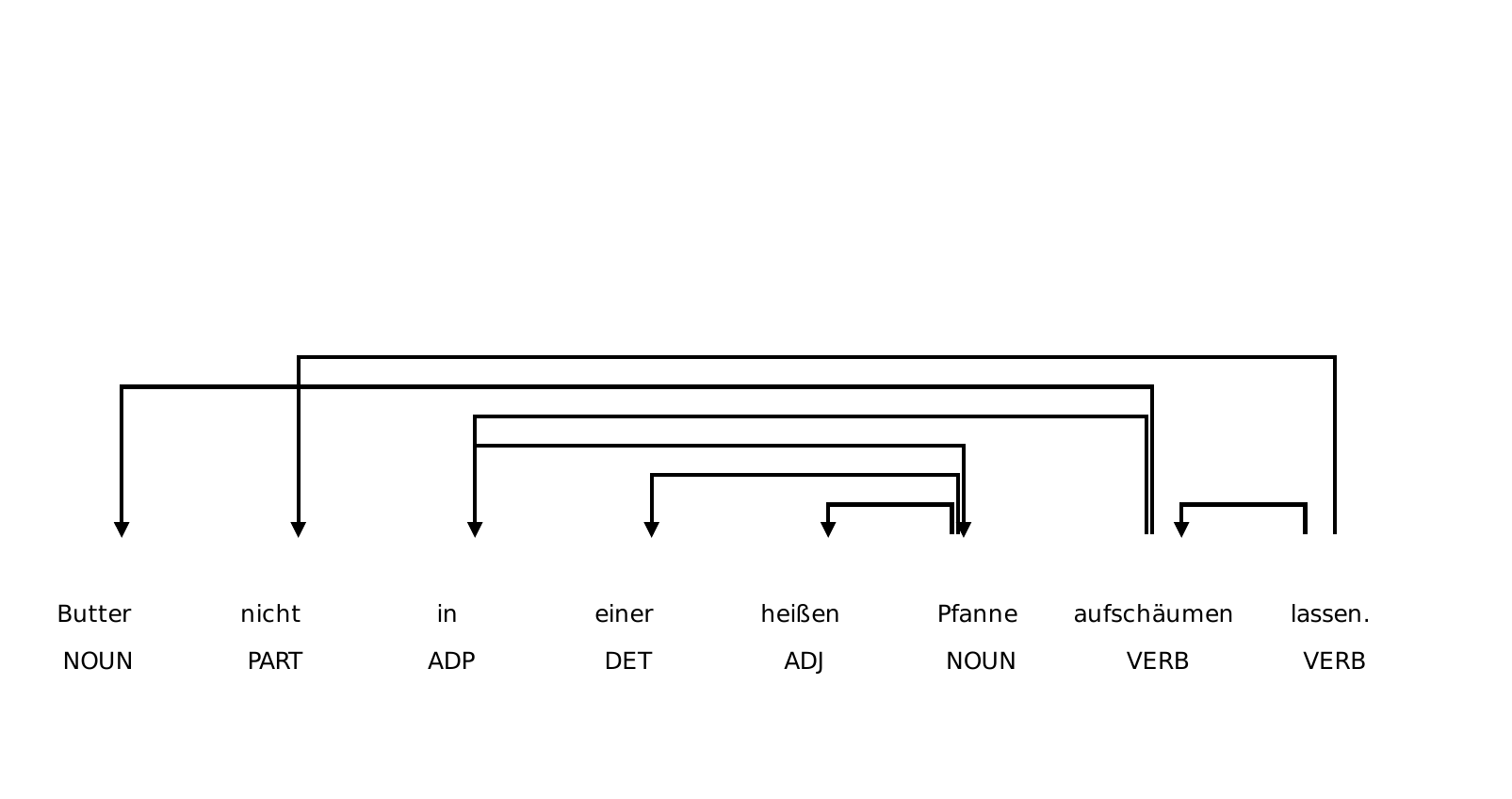}
    \caption{Dependency tree of a negated German recipe sentence}
    \label{fig:neg_sentence}
\end{figure}

\autoref{fig:dp_sentence} shows the exemplary dependency tree of a sentence $S=$ \textit{"Butter in einer heißen Pfanne aufschäumen lassen."} (Engl.: \textit{"Foam butter in a hot pan."}) of a text $\txt = (S)$. By traversing the dependency graph for all verbs in $S$ we obtain the activity set $\acts{\txt} = \{(v, s, o, m)\}$ with:
\begin{itemize}
    \item $v = \{\text{aufschäumen, lassen}\}$
    \item $s = \emptyset$
    \item $o = \{\text{Butter}\}$
    \item $m =\{\text{in einer heißen Pfanne}\}$
\end{itemize}
\paragraph{Negations.}
The negation of an activity constitutes a special case. \autoref{fig:neg_sentence} illustrates the dependency tree of $S$ with negotiation. The dependency parser tags negation dependencies as $ng$ and thus, allows us to extract negations \citep{aa2019extracting,albert2003tiger}. We omit negations in our extraction approach. However, negations could easily be added to the activity if needed.

\paragraph{Quantification}
Not every activity described in a text is mandatory. While constraint-markers, as declared by \citep{aa2019extracting,winter2019deriving,winter2018detecting}, do not suffice for the generic classification of sentences containing process information, they indicate, if present, whether there \textit{exists} a path in the supposed workflow net $\wf$ of a text $\txt$ containing a related activity $a$ or if \textit{all} paths of the workflow net contain $a$. We used GermaNet \citep{hamp1997germanet} to obtain a more complete list of constraint markers as given in \autoref{tab:markers}.  
\begin{table}[htb]
    \centering
    \begin{tabular}{|l||p{5cm}|}
    \hline 
        Marker & Word \\ \hline
         \textit{EXISTS} & können, dürfen, mögen, sollten, kann, vielleicht, optional,
                  eventuell, gegebenenfalls \\\hline
         \textit{ALL} & müssen \\\hline
    \end{tabular}
    \caption{Quantifying constraint markers}
    \label{tab:markers}
\end{table}
By default, if not further specified, we assume an activity to be mandatory.

\paragraph{Irrelevant Subordinate Clauses.}
Although we may assume to extract activities from relevant sentences only, we may not assume every sentence's verb to be relevant. For instance, the sentence $S=$ \textit{Butter in einer heißen Pfanne aufschäumen lassen, das schmeckt mir am besten} contains the relation $a_1 = a$ as in \autoref{fig:dp_sentence}. However, simply extracting all verbs and their dependents would also yield $a_2 = (\{\text{schmeckt}\}, \{\text{das}\}, \emptyset, \{\text{am besten}\})$. A simple heuristic to handle such clauses is to use the VVIMP tag from \citep{albert2003tiger} as incorporated into the spaCy framework. However, as recipes are not formalized, some are written in a descriptive form or a first-person narrative. Therefore, such recipes would not be handled well. A second heuristic may be the recognition of a switch in writing style. If a sentence contains an imperative and a non-imperative verb, we may assume the imperative verb to be an activity and the non-imperative to be descriptive. The effect of the heuristic is examined in \autoref{sec:results}.

\subsection{Activity Ordering: Interactivity Relation Extraction}
By default, we implicitly assume the described activities in the process model to be ordered as their appearance in the text orders them.
However, interactivity relations explicitly describe the activity ordering and can be classified as \textit{AND}, \textit{OR}, or \textit{BEFORE} relations. To obtain the order in which the activities described in the text should be executed, we need to be able to extract these interactivity relations. In the simpler case, these relations are expressed within a sentence. Coordinating conjuncts in combination with synonym databases such as WordNet \citep{miller1998wordnet} or the German GermaNet \citep{hamp1997germanet} as tagged by the dependency parser can be used to identify conjunctions and disjunctions of activities to extract  \textit{AND} or \textit{OR} relations. Temporal adverbs can be identified using WordNet/GermaNet as well (cf. \citep{aa2019extracting}).
\textit{BEFORE} relations that are described across sentences can be handled using coreference resolution to identify the referenced activities from previous sentences, or using simple heuristics. For instance, it is reasonable to assume that a temporal adverb as \textit{inzwischen} (Engl. meanwhile) indicates an \textit{AND} relation to the activities of the previous sentence. 
In sum, we identified the following heuristics:
\begin{itemize}
    \item coordinating conjuncts within sentences
    \item temporal adverbs within sentences (if not dependent on the first activity):
    \begin{enumerate}
        \item if indicating \textit{AND} relation: add \textit{AND} relation to previous activity
        \item if indicating \textit{BEFORE} relation: add \textit{BEFORE} relation to activities in the previous sentence
    \end{enumerate}
    \item temporal adverbs across sentences (if dependent on the first activity in the sentence): 
    \begin{enumerate}
        \item if indicating \textit{AND} relation: add \textit{AND} relation to activities of previous sentence
        \item if indicating \textit{BEFORE} relation and only one activity within sentence: add \textit{BEFORE} relation to activities in the previous sentence
    \end{enumerate}
\end{itemize}
The indicator synonyms are given in \autoref{tab:temps}

\begin{table}[htb]
    \centering
    \begin{tabular}{|l||p{5cm}|}
    \hline 
        Adverb & Adverb \\\hline
         \textit{BEFORE} & zuvor, davor, vorab, vordem, vorher, vorweg, zuerst, zunächst, anfänglich, anfangs, eingangs, erst, vorerst  \\\hline
         \textit{AND} & inzwischen, dabei, währenddessen, dazwischen, inzwischen, mittlerweile, solange, zwischenzeitlich, derweil, einstweilen\\\hline
    \end{tabular}
    \caption{Temporal Adverbs for the extraction of interactivity relations}
    \label{tab:temps}
\end{table}
\subsection{\uppercase{Generating Process Models}}
\begin{figure}[htb]
    \centering
    \subfigure[]{\includegraphics[width=0.24\linewidth]{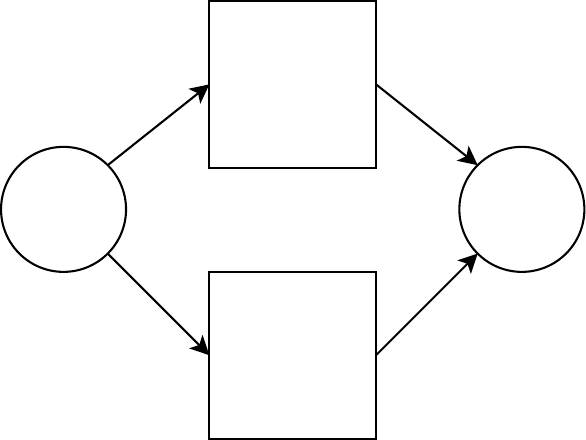}} 
    \subfigure[]{\includegraphics[width=0.66\linewidth]{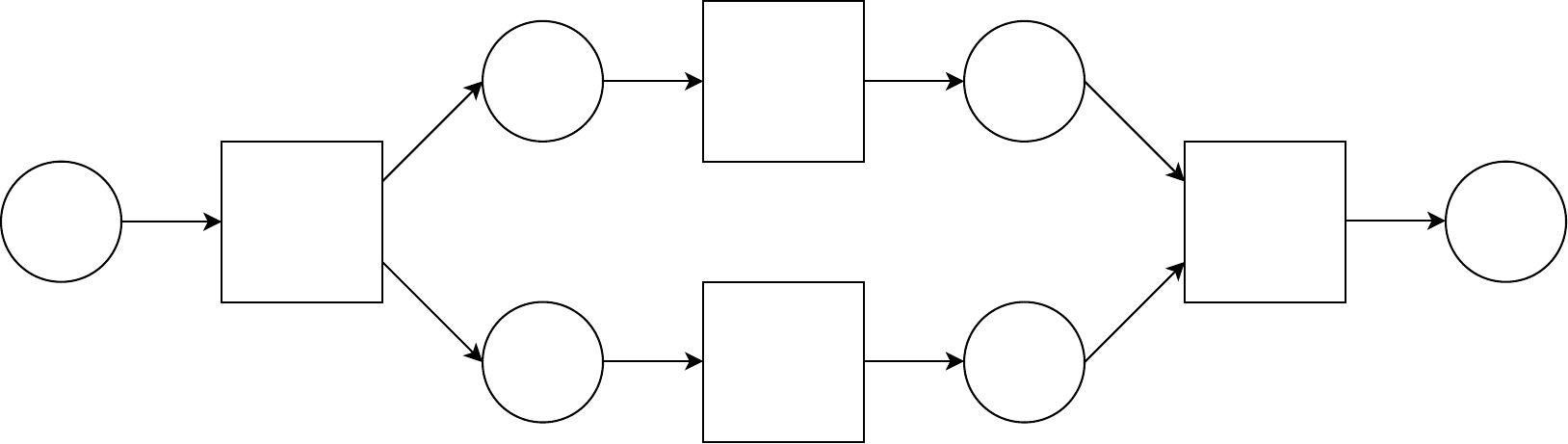}} 
    \caption{(a) OR pattern (b) AND pattern}
    \label{fig:patterns}
\end{figure}

From the previous steps, we obtain a set of activities and a set of binary relations between activities. The remaining task is the creation of a workflow net. To do so, we first, create a workflow net for each sentence by applying patterns (cf. \autoref{fig:patterns}) for OR, AND, and BEFORE relations extracted as described in \autoref{subsub:rel}. 

Then, the sub nets are merged to the final workflow net $\wf$ of the recipe $\txt$ by either appending the sub net to the previous sub net or, if the the first activity in a sentence indicates a parallelization, the sub net is added using AND pattern as a parallel to the previous sub net (cf. Algorithm \ref{alg:wfg}). 

\begin{algorithm}
\caption{Workflow net generation}\label{alg:wfg}
\begin{algorithmic}[1]
\Function{GenerateWorkflowNet}{$\txt$}
    \State $pn := NewPetriNet()$
    \State $last\_sn := pn$
    \State $N := len(\txt)$
    \For{$i \in \{1,\dots, N\}$}
        \State $sn := get\_sub\_net(\acts{S_i})$
        \If{$parrallel(S_i)$}
            \State $pn.add\_parallel(last\_sn, sn)$
        \Else
            \State $pn.append(sn)$
        \EndIf
        \State $last\_sn := sn$
    \EndFor
\EndFunction
\end{algorithmic}
\end{algorithm}

\section{\uppercase{Experiments}}
\label{sec:experiments}
Rule-based and machine-learning-based approaches to PME formulate a trade-off. While rule-based approaches require the adoption of rules to suit domain-specific formulations and conventions, machine-learning-based approaches require large corpora of complexly annotated data. Thus, as formulated by e.g. \citep{qian2020approach}, we may divide PME into different tasks to be solved either machine-learning-based or rule-based.

\subsection{\uppercase{Data \& Data Preparation}}
Recipes from the German recipe website Chefkoch\footnote{\url{https://www.kaggle.com/datasets/sterby/german-recipes-dataset}} were used to train the sentence classifiers and evaluate GUIDO. The dataset contains 44672 unique sentences from 4291 recipes, from which we sub-sampled and annotated 2030 recipes for binary classification and 50 mutually exclusive recipes for workflow net annotation, to compare the extracted process model to. 

For the sake of training the BERT text classifier, we identified and replaced URLs by a unique $\$URL$ token, using regular expressions. The rule-based PME levels do not require further text normalization.

\begin{table}[htb]
\centering
\begin{tabular}{ |l||c|c|c|c| } 
\hline
Set & \# S & \% S & \# Relevant & \% Relevant\\
\hline
Train & 1533 & 60\% & 773 & $50.42$\% \\\hline
Dev & 512 & 20\% &  240 & $46.86$\%\\\hline
Test & 511 & 20\% & 265 & $51.75$\%\\ %
\hline
\end{tabular}
\caption{Sentence corpus statistics where S denotes Sentences after balancing by down-sampling}
\label{tab:sent}
\end{table}

\paragraph{Sentence \& Workflow net Labeling.}
A sentence dataset was build using the spaCy dependency-parser-based sentence tokenizer \citep{honnibal2020spacy}. Two annotators labeled the sentences. To increase the process quality of the labeling process and increase the quality of the labeled dataset, labeling guidelines were written before labeling\footnote{cf. \url{https://github.com/nils-freyer/GUIDO/wiki/Labeling-Guideline}}. If there was uncertainty in assigning a label in a given sentence, the annotator discussed the label with the other annotator and updated the labeling guidelines with the result of the discussion. Subsequently to the sentence annotation process, the sentences were further sub-sampled to obtain a balanced dataset of $3150$ annotated sentences, as irrelevant sentence make about $10\%$ of the sentence population only. The sub-sampled sentence corpus was split into \textit{train}, \textit{test} and \textit{dev} sets for training and evaluation. The statistics of the annotated sentence corpus are given in \autoref{tab:sent}.
A set of $50$ recipes with $616$ sentences in total was annotated with corresponding workflow nets by a single annotator.

\subsection{\uppercase{Evaluation}}
To evaluate the performance GUIDO, the text classification and the PME task are evaluated separately. The text classification task was evaluated according to its \textit{F1-Score} on a validation set of size $N=512$. A total of $50$ recipes were annotated manually using ProM\footnote{\url{https://promtools.org/}}, in order to obtain similarity metrics. As, in the case of PME, we need a metric that compares the behavior of workflow nets rather than the syntactical equivalence of the output to the annotation, we implemented a behavioral similarity score based on causal footprints, an abstract representation of a Petri net's behavior. \citep{mendling2007degree}. We applied the similarity metric to a rule-based baseline model, GUIDO with heuristics to handle subordinate clauses and GUIDO without additional heuristics.
All experiments were done on using a single machine with an Intel Xeon processor, a NVIDIA GeForce RTX-A5000 GPU with 16 GB of VRAM, and 64 GB of RAM, running on Ubuntu 20\.04, which has a estimated carbon efficiency of 0.432 kgCO$_2$eq/kWh. A cumulative of 0.5 hours of computation was performed on hardware of type RTX A5000 (TDP of 230W) for training. A cumulative of 30 hours of computation was performed on hardware of type Intel Xeon W-11855M (TDP of 45W) for evaluating.
Total emissions are estimated to be 0.65 kgCO$_2$eq of which 0 percents were directly offset. Estimations were conducted using the \href{https://mlco2.github.io/impact#compute}{MachineLearning Impact calculator} presented in \cite{lacoste2019quantifying}.

To conduct our experiments, we fully parameterized the project using a hydra-config\footnote{\url{https://hydra.cc}}. A parallelized grid search was used for parameter tuning. Furthermore, we used the mlflow framework\footnote{\url{https://mlflow.org}} for visualizing training and evaluation metrics.

\section{\uppercase{Results}}
\label{sec:results}
In this section, we first compare the proposed BERT sentence classifier with three baseline models, evaluated on $512$ unseen sentences. Then, we evaluate GUIDO on $50$ unseen recipes, containing 616 sentences.

\paragraph{Sentence Classification.}
Multiple approaches were evaluated in addition to the BERT sentence classifier and compared to the VVIMP baseline (cf. \autoref{tab:scores}). 
\begin{table}[t]
    \centering
    \begin{tabular}{|l||p{1cm}|p{0.8cm}|p{0.8cm}|p{0.8cm}|p{0.8cm}|}
    \hline 
         Score & VVIMP & Log Reg & LSTM FT & LSTM GloVe & \textbf{BERT}  \\\hline
         F1 & 0.58 & 0.90 & 0.91 & 0.92 & \textbf{0.973}\\\hline
    \end{tabular}
    \caption{Classifier F1-Scores for the base line heuristic (VVIMP), the logistic regression classifier (Log Reg), the LSTM classifier with FastText (LSTM FT), the LSTM classifier with GloVe (LSTM GloVe) and the BERT sentence classifier.}
    \label{tab:scores}
\end{table}
The simple VVIMP heuristics classifies a sentence as process relevant, i.e., containing at least one activity, if there is no subject that is not a child of an imperative in the dependency tree, resulting in an F1-Score of $\approx 0.81$.
Further, the documents were tfidf-vectorized. A binary logistic regression classifier was trained and obtained an F1-Score of $\approx 0.90$. A simple LSTM with a text-classification head obtained an F1-Score of $\approx 0.91$ on fine-tuned GloVe vectors and $\approx 0.92$ on pre-trained multilingual fasttext vector. 
Finally, the BERT sentence classifier outperformed the baseline models with a final F1-Score of $\approx 0.973$ with batch size $16$, $5$ epochs and learning rate $3e^{-5}$.

\paragraph{Process Model Extraction.}
We compared the $50$ annotated workflow nets to the extracted workflow nets by GUIDO + VVIMP heuristic, GUIDO - VVIMP heuristic, and to the extracted workflow nets of a purely rule-based approach.
The results (cf. \autoref{tab:pme}) show significant improvements for the rule-based process extractor when adding the text classification level with an average similarity score of $\approx 0.93$ over $\approx 0.84$. The usage of a VVIMP heuristic to handle subordinate clauses did not have a significant effect on the performance of GUIDO, as only one verb was classified as an imperative by the tagger.

\begin{table}[ht]
    \centering
    \begin{tabular}{|l||p{1.3cm}|p{1.7cm}|p{1.7cm}|}
    \hline 
         Model & Rule-Based & GUIDO\ - VVIMP & \textbf{GUDIO + VVIMP}  \\\hline
         CFP-Sim & 0.84 & 0.93 & \textbf{0.93}\\\hline
    \end{tabular}
    \caption{CFP behavioral similarities}
    \label{tab:pme}
\end{table}

\section{\uppercase{Discussion \& Future Work}}

The proposed PME model GUIDO shows good performance given a reduced labeling effort of $2030$ binary annotated sentences compared to purely machine-learning-based approaches. The additional step of a sentence classifier significantly improves the performance of rule-based PME models compared to purely rule-based PME models and therefore, formulates a compromise to the annotation cost and specificity trade-off. The rule-based level of GUIDO was designed in a generic way, applicable to multiple domains. Additionally, the approach is easily transferable for rule-based Declarative PME tasks \citep{aa2019extracting,lopez_assisted_2019}.

The most common errors of GUIDO were miss classifications of sentences and irrelevant subordinate clauses. Common taggers perform poorly on process data, as they were mostly trained on news data \citep{han2019novel}. In particular, news data rarely contains imperatives and thus has a high miss classification rate for the VVIMP tags we use for handling irrelevant subordinate clauses. A further limitation to the PME task and results is the fine graindness of the desired process model. Throughout this paper, we assumed repetitive activities to be a single activity. For instance \textit{"wiederholt umrühren"} (Engl.: \textit{"stir repeatedly"}) would not result in a cycle in the Petri net but be a single transition. Such cycles should be incorporated and evaluated in future work. 
GUIDO was trained and applied to German recipes only in this paper, containing imperative process models only. In future work, we will evaluate our approach on declarative guidelines. While there is a lack of comparison for German workflow net extraction, the behavioral similarity scores achieved by GUIDO seem competitive to related work in other languages \citep{qian2020approach}. Especially the rule-based level of GUIDO is grammar specific, which is specific to the German language. Thus, we will adopt and evaluate GUIDO on English recipes in the future to get further insights on it performance compared to existing state of the art approaches.

\section*{\uppercase{Acknowledgments}}
This research has been developed and funded by the project Assist.me (grant number 16KN090726) of the German Federal Ministry of Economic Affairs and Climate Action (Bundesministerium für Wirtschaft und Klimaschutz (BMWK)).
\bibliographystyle{apalike}
{\small
\bibliography{main}}

\end{document}